\documentclass[10pt,twocolumn,letterpaper]{article}
\usepackage{cvpr}

\usepackage{times}
\usepackage{epsfig}
\usepackage{graphicx}
\usepackage{amsmath}
\usepackage{amssymb}
\usepackage{mathtools}
\usepackage{microtype}
\usepackage[inline]{enumitem}
\usepackage{nicefrac}
\usepackage[group-separator={,}]{siunitx}
\usepackage[numbers,sort]{natbib}
\usepackage{booktabs}
\usepackage[dvipsnames]{xcolor}
\usepackage{float}
\usepackage[symbol]{footmisc}

\setlist{leftmargin=*,noitemsep}

\usepackage[breaklinks=true,colorlinks,bookmarks=false,pagebackref=true]{hyperref}
\usepackage[capitalize]{cleveref}
\crefname{section}{\S}{\S}
\Crefname{section}{Section}{Sections}
\Crefname{table}{Table}{Tables}
\crefname{table}{Tab.}{Tabs.}

\newcommand{\R}{\mathbb{R}}
\newcommand{\E}{\mathbb{E}}
\newcommand{\LL}{\mathcal{L}}
\newcommand{\N}{\mathcal{N}}
\newcommand{\U}{\mathcal{U}}

\DeclareMathOperator*{\argmin}{arg\,min}
\DeclareMathOperator{\proj}{proj}
\newcommand{\dext}{\mathrm{d}}

\begin{document}

\title{DeepCurrents: Learning Implicit Representations of Shapes with Boundaries}

\author{David Palmer$^{*1}$ \quad Dmitriy Smirnov$^{*1}$ \quad Stephanie Wang$^2$
\quad Albert Chern$^2$ \quad Justin Solomon$^1$\\
$^1$Massachusetts Institute of Technology \quad $^2$UC San Diego 
}

\maketitle

\renewcommand{\thefootnote}{\fnsymbol{footnote}}
\footnotetext[1]{Authors contributed equally to this work.}

\begin{abstract}
Recent techniques have been successful in reconstructing surfaces as level sets of learned functions (such as signed distance fields) parameterized by deep neural networks. Many of these methods, however, learn only closed surfaces and are unable to reconstruct shapes with boundary curves. We propose a hybrid shape representation that combines explicit boundary curves with implicit learned interiors. Using machinery from geometric measure theory, we parameterize currents using deep networks and use stochastic gradient descent to solve a minimal surface problem. By modifying the metric according to target geometry coming, e.g., from a mesh or point cloud, we can use this approach to represent arbitrary surfaces, learning implicitly defined shapes with explicitly defined boundary curves. We further demonstrate learning families of shapes jointly parameterized by boundary curves and latent codes.
 \end{abstract}

\section{Introduction}

Shape representation is a crucial component of geometry processing and learning algorithms. Depending on the target application, different representations have varying tradeoffs. Broadly, shape representations fall naturally into two classes: \emph{Lagrangian} or \emph{explicit}; \emph{Eulerian} or \emph{implicit}. In this work, we show how to use the theory of \emph{currents} from geometric measure theory to design a flexible neural representation that combines favorable aspects from each category, representing the interiors of surfaces implicitly while maintaining an explicit representation of their boundaries.

Lagrangian representations encode a shape by giving coordinates of points or parameterizing regions of the shape. To represent a curve in a Lagrangian way, one might give coordinates of successive points along the curve. 
Analogously, to represent a surface in 3D, one might use a mesh, which assembles the surface out of simple patches. 
Lagrangian representations afford great precision but require predetermined combinatorial structures, making it difficult to represent families of shapes with varying topology. 

In contrast, Eulerian representations encode a shape via a function on some background domain. For example, a surface might be encoded as the \emph{level set} of a scalar function sampled on a regular grid. Level sets of signed distance fields (SDFs) 
form one popular implicit representation. Implicit functions naturally capture topological variation, but traditional implicit shape representations, in which the background geometry must be discretized with a fixed grid or mesh, waste resolution on regions far away from the level set of interest. Recent \emph{neural implicit representations} alleviate this problem \cite{chen2019learning,OccupancyNetworks,Park_2019_CVPR}. The universal approximation and differentiability properties of neural networks make them an appealing alternative to regular grid discretizations.

Neural implicit representations come with their own limitations. Like other implicit representations based on level sets, most neural implicit representations can only encode closed surfaces, which lack boundary curves. Boundaries are desirable as they can provide manipulation handles for controllable deformation, and common boundaries can be used to stitch together surfaces into a larger articulated surface.

In this paper, we describe a new way to encode neural implicit surfaces with boundaries, which can then be combined into more complex hybrid surfaces.
The key to our representation is the theory of currents from geometric measure theory. In this theory, $k$-dimensional submanifolds are defined by their integration against differential $k$-forms, generalizing how distributions ($0$-currents) are defined by integration against smooth functions. Current spaces are complete normed linear spaces that make optimization over surfaces convenient, and the boundary operator also becomes linear on these spaces. Classically, currents were the key to solving Plateau's \emph{minimal surface} problem by transforming it into \emph{mass norm minimization}. We adopt the mass norm as the primary loss function encouraging our neural currents to converge to smooth surfaces.

We demonstrate our representation with three applications. We first demonstrate how it enables computing minimal surfaces efficiently through stochastic gradient descent. Then, by modifying the background metric used to define the mass norm, we reconstruct arbitrary surfaces from data. Finally, we demonstrate the flexibility of our representation by encoding families of surfaces with explicit boundary control.

\paragraph*{Contributions.}
In summary, we
\begin{itemize}
    \item propose a new neural implicit surface representation with explicit boundary curves;
    \item show how to use SGD on the mass norm to compute minimal surfaces;
    \item introduce a custom background metric and additional loss terms to represent surfaces from data; and
    \item describe a framework for learning families of surfaces parameterized by their boundaries along with a latent code.
\end{itemize}

\section{Related Work}

Our work takes classical ideas in minimal surface computation and brings them into the context of modern deep learning to form a new neural shape representation. Below, we summarize key prior works in these two areas.

\subsection{Minimal Surface Computation}
Most computational approaches to minimal surface generation use a mesh or grid representation of the surface. 
In the twentieth century, numerical minimal surface problems were discretized by finite difference methods on a grid, assuming the surfaces were function graphs \cite{Douglas:1927:MNS, Concus:1967:NSM}.  Grid-based methods were 
later adapted to
triangle meshes \cite{Wilson:1961:DDP,Hinata:1974:NSP}, allowing the generated surface to leave the space of function graphs \cite{Wagner:1977:CNA}. Modern mesh-based minimal surface solvers use mean curvature flow \cite{Dziuk:1990:AES,Brakke:1992:SE,Desbrun:1999:IF}, stretched grids \cite{Popov:1996:VFM}, quasi-Newton iterations \cite{Pinkall:1993:CDM,Schumacher:2019:VCD}, Voronoi tessellations \cite{pan2012robust}, or curvature flows with a conformal constraint \cite{Crane:2011:STD,Kazhdan:2012:MCF}.  These methods based on explicit surface representations are straightforward, but the optimization often suffers from local minima due to the non-convexity of the area functional and can even diverge if the initial mesh has the wrong topology \cite{Wagner:1977:CNA, Pinkall:1993:CDM}.

A different approach to the minimal surface problem is based on \emph{geometric measure theory} (GMT), whose theoretical foundations were developed in the 1960s \cite{Federer:1960:NIC,Morgan:2016:GMT,Fleming:2015:GMT}.  
In this theory, curves and surfaces are represented implicitly by \emph{currents} as dual to differential forms.
Such representations have been used in geometry processing \cite{Mullen:2007:EGP,Buet:2018:DAS,Mollenhoff:2019:LVV,mollenhoff2019flat} and medical imaging \cite{Charon:2014:FC, Vaillant:2005:SMVC, Glaunes:2004:DMD, Durrleman:2008:DDC, Durrleman:2009:SMC, Durrleman:2011:RAEVA}.
In geometric measure theory, the minimal surface problem becomes the convex \emph{minimal mass norm} problem (see \Cref{sec:minmass}). A discrete analog of the minimal mass problem on a graph is a linear program known as the optimal homologous chain problem \cite{Sullivan:1990:CAT,Dunfield:2011:LSA,Dey:2011:OHC,Cohen:2020:LOHC}.  
GMT-based discretization of the minimal surface problem in Euclidean space was pioneered by \cite{Parks:1997:CLH} and revisited by \cite{Brezis:2019:PPP,wang2021computing}.

\subsection{Deep Learning for Shape Reconstruction}

Using deep learning to produce 3D geometry has gained popularity in vision and graphics. Network architectures now can output many explicit shape representations, like voxel grids \cite{delanoy20183d,genre,wu2018learning}, point clouds \cite{fan2017point,yin2018p2p,pointflow}, meshes \cite{nash2020polygen,wang2018pixel2mesh,Hanocka2020p2m}, and parametric primitives \cite{sharma2020parsenet,abstractionTulsiani17,Paschalidou2019CVPR,smirnov2020dps,smirnov2021patchbased}. While the content produced by these approaches is generally easy to render and manipulate, it is often restricted in topology and/or resolution, limiting expressiveness.

A different approach circumvents topology and resolution issues by representing 3D shapes \emph{implicitly}, using functions parameterized by neural networks. In DeepSDF, \citet{Park_2019_CVPR} learn a field that approximates signed distance to the target geometry, while \citet{OccupancyNetworks} and \citet{chen2019learning} classify query points as being outside or inside a shape. Others further improve the results by proposing novel regularizers, loss functions, and training or rendering approaches \cite{amos2020,Atzmon_2020_CVPR,takikawa2021nglod,lipman2021phase}. While these works achieve impressive levels of detail in surface reconstruction, they largely suffer from two drawbacks---lack of control and inability to represent open surfaces, i.e., those with boundary.

Neural implicit learning methods typically overfit to a single target shape or learn a family of shapes parameterized by a high-dimensional latent space. While recent work has shown the possibility of adapting classical geometry processing algorithms to neural implicit geometries \cite{Yang_2021_NeurIPS}, applying targeted manipulations and deformations to learned shapes remains nontrivial. Several papers propose \emph{hybrid representations}, combining the expressive power of neural implicit representations with the control afforded by explicit representations. \citet{genova2020local} reconstruct shapes by learning multiple implicit representations arranged according to a learned template configuration. In DualSDF \cite{zekun2020dualsdf}, manipulations can be applied to learned implicit shapes by making changes to corresponding explicit geometric primitives. BSP-Net \cite{chen2020bspnet} and CvxNet \cite{deng2020cvxnet} restrict the class of learned implicit surfaces to half-spaces and convex hulls, respectively. \cite{Liu2021MLS} defines local implicit functions on point clouds, facilitating the transition between discrete points and smooth surfaces during training.

Because neural implicit shapes are typically level sets of learned functions, this limits the class of representable shapes to closed surfaces. Two notable exceptions are \cite{chibane2020ndf}, which learns \emph{unsigned} distance functions rather than SDFs, and \cite{Venkatesh_2021_ICCV}, which maps an input point to its closest point on the target surface. Our DeepCurrents adopt a hybrid representation, which models boundaries explicitly and allows them to be used as handles for manipulation.
\section{Preliminaries}

Geometric measure theory is a vast field
that we will not attempt to summarize here. For a comprehensive treatment, we refer the reader to \cite{simon2014introduction, federer-gmt, lang2005}. We focus on the rudiments necessary to construct our optimization problem in dimensions two and three, eliding technical issues that arise in higher-dimensional ambient spaces.

The theory of currents is motivated by solving \emph{Plateau's Problem}, the problem of finding the surface of minimal area enclosed by a given boundary:
\begin{equation}
    \argmin_{\Sigma} \{A(\Sigma) : \partial \Sigma = \Gamma\}. \label{eq:plateau}
\end{equation}
The problem \eqref{eq:plateau} seeks a solution in the space of smooth embedded submanifolds with boundary, which lacks convenient properties such as convexity and compactness required for reasoning about optimization. In GMT, this space is \emph{relaxed} to a space of \emph{currents}, generalized submanifolds characterized by integration. Plateau's problem \eqref{eq:plateau} is systematically translated into a problem over currents. The area functional becomes the convex \emph{mass norm}, and $\partial$ becomes a linear operator constructed by dualizing the exterior derivative $\dext$. We describe this translation in detail below.

\subsection{Currents}

Currents are to submanifolds as distributions are to sets of points. Just as distributions are characterized by integration against functions, $k$-currents are characterized by integration against differential $k$-forms in the ambient space. 
For our purposes, that ambient space will be an open subset $U \subseteq \R^d$, $d \le 3$. For now, we will assume the metric is Euclidean; see \Cref{sec:metric} for the generalization
to Riemannian metrics. We will also assume that $U$ is bounded and contractible to elide various technical issues.

We denote the space of smooth $k$-forms with compact support in $U$ by $\Omega_c^k(U)$. Recall that a $k$-form $\zeta \in \Omega_c^k(U)$ smoothly assigns to each point $x\in U$ an element $\zeta_x \in \bigwedge^kT_x^*U$, the exterior power of the cotangent space at $x$. In Euclidean space, there is a canonical identification between covectors ($k=1$) and vectors. A Riemannian metric provides a similar identification but requires more careful bookkeeping (see \Cref{sec:metric}).

The space of $k$-currents
\begin{equation} \mathcal{D}_k(U) = (\Omega_c^k(U))^* \end{equation}
is the \emph{dual space} of (compactly-supported) $k$-forms, i.e., it consists of continuous linear functionals on $k$-forms. An element $T \in \mathcal{D}_k(U)$ is defined by its assignment of real values to $k$-forms:
\begin{equation} \zeta \in \Omega_c^k(U) \mapsto T(\zeta) \in \R. \end{equation}

The following are two key examples of currents:
\begin{itemize}
    \item A $0$-current is simply a distribution, as
    \begin{equation} \mathcal{D}_0(U) = (\Omega_c^0(U))^* = (C_c^\infty(U))^* = \mathcal{D}(U). \end{equation}
    \item A submanifold $\Sigma \subset U$ of dimension $k$ can be viewed as a current $[\Sigma] \in \mathcal{D}_k(U)$ by integration against it:
    \begin{equation} [\Sigma](\zeta) \coloneqq \int_\Sigma \zeta. \end{equation}
\end{itemize}

\subsection{Boundary Operator}
In generalizing the boundary operator from submanifolds to currents, we need to ensure that $\partial[\Sigma] = [\partial \Sigma]$. Stokes' Theorem tells us that
\begin{equation} [\partial \Sigma](\zeta) = \int_{\partial \Sigma} \zeta = \int_\Sigma \dext\zeta = [\Sigma](\dext\zeta), \end{equation}
motivating the definition
\begin{equation} \partial T(\zeta) \coloneqq T(\dext\zeta). \end{equation}
In words, we
define $\partial$ as the adjoint of $\dext$.

\subsection{Mass Norm}
As we did for the boundary operator, we write the area functional in terms of integration against forms and then replace integration by current evaluation. This definition depends on a pointwise norm $|\cdot|$ on $k$-forms.
As we are working in dimensions $d \le 3$, it is sufficient to use the pointwise inner product norm, and we will not concern ourselves with complications that occur in higher dimensions.

If $\Sigma \subset U$ is a smooth $k$-submanifold with boundary, then its area satisfies
\begin{equation}
    A(\Sigma) = \sup_{\zeta \in \Omega_c^k(U)} \left\{ \int_\Sigma \zeta : |\zeta_x| \le 1 \; \forall x \in U \right\}.
\end{equation}
So we define the \emph{mass norm} of a
current $T \in \mathcal{D}_k(U)$ as:
\begin{equation} \mathbf{M}(T) \coloneqq \sup_{\zeta \in \Omega_c^k(U)} \{ T(\zeta) : |\zeta_x| \le 1 \; \forall x \in U \}. \end{equation}

\subsection{Minimal Mass Problem} \label{sec:minmass}
Applying the transformations above to the problem \eqref{eq:plateau}, one obtains a relaxation known as the \emph{minimal mass problem}:
\begin{equation}
    \min_{T \in \mathcal{D}_k(U)} \{ \mathbf{M}(T) : \partial T = \Gamma \}.
    \label{eq:massmin}
\end{equation}
In classical GMT, the current $T$ is taken to be in the space $\mathcal{I}_k(U)$ of \emph{integral $k$-currents}, which roughly means currents that look like integer linear combinations of Lipschitz surfaces. When optimizing over $\mathcal{I}_{d-1}(U)$ in ambient dimension $d \le 7$, there is an optimal solution corresponding to a smooth submanifold (see \cite{federer-gmt} Theorem 5.4.15, \cite{simon2014introduction} Theorem 5.8, \cite{lang2005} Theorem 3.10). More recent theory extends this result to optimization over general currents $\mathcal{D}_k(U)$ (see \cite{Brezis:2019:PPP} Theorem 2, \cite{simon2014introduction} Remark 5.2).

\subsection{Representing Currents by Forms}
For computational purposes, we follow \cite{wang2021computing} and optimize over $k$-currents represented by differential $(d-k)$-forms.
This allows us to represent currents by neural networks.

A $(d-k)$-form can be identified with a $k$-current $[\omega] \in \mathcal{D}_k(U)$ by defining
\[[\omega](\zeta) \coloneqq \int_U \omega \wedge \zeta \]
for any $\zeta \in \Omega_c^k(U)$. With this identification, we have by Stokes' Theorem:
\begin{align}
    \partial[\omega](\zeta) &= [\omega](\dext\zeta) = \int_U \omega \wedge \dext\zeta \\
    &= \nonumber (-1)^{d - k + 1}\int_U \dext\omega \wedge \zeta = \left[(-1)^{d - k + 1} \dext\omega\right](\zeta).
\end{align}
The boundary constraint $\partial [\omega] = \Gamma$ thus becomes an exterior differential equation,
\begin{equation}
    \dext\omega = \delta_{\Gamma}, \label{eq:bdry-de}
\end{equation}
where $\delta_{\Gamma}$ is a singular $(d-k)$-form representing $\Gamma$. 

Similarly, the mass norm of a $k$-current becomes the $L^1$ norm of a $(d-k)$-form:
\begin{equation}
    \mathbf{M}([\omega]) = \|\omega\|_1 = \int_U |\omega(x)| \dext\mathrm{vol}
    \label{eq:mass-l1}
\end{equation}
where $\dext\mathrm{vol}$ is the volume form.

\section{DeepCurrents}

In the previous section, we described the relaxation of Plateau's minimal surface problem into a convex optimization problem over a space of currents, with the property that its optima include smooth surfaces, and we showed how to represent certain currents by differential forms. In the following section, we introduce our novel neural representation of currents and SGD mass minimization.

\subsection{Neural Representation}
The linear space of solutions to \eqref{eq:bdry-de} can be parameterized using the Hodge decomposition as follows:
\begin{equation}
\omega = \dext f + \alpha, \label{eq:alpha-hodge}
\end{equation}
where $\alpha \in \Omega^{d-k}(U)$ is any particular solution of \eqref{eq:bdry-de}, $f \in \Omega^{d-k-1}(U)$; we ignore the harmonic term as $U$ is contractible. For curves in $U \subset \R^2$ ($k = 1$, $d = 2$) and surfaces in $U \subset \R^3$ ($k=2$, $d=3$), $f$ will simply be a function on $U$, which we can represent by a neural network. It is convenient to use the (Euclidean) musical isomorphism $\sharp$ to encode our $1$-form $\omega$ as the vector field $\omega^\sharp$. Intuitively, the vector field corresponding to a current points in the surface normal direction. Under this identification, $(\dext f)^\sharp = \nabla f$, which can be computed by autodifferentiation.

As for $\alpha$, there is a particularly convenient choice known as the \emph{Biot-Savart field}, which can be written in closed form when $\Gamma$ is a polygonal curve (see \cite{prautzsch2009}):
\begin{equation}
    \alpha^{\sharp}(x) \! \coloneqq \! \int_{\Gamma} \frac{d\vec{\ell} \times \vec{r}}{|\vec{r}|^3} \! = \! \sum_i \frac{(\hat{t}_i \cdot (\hat{r}^1_i - \hat{r}^0_i))(\hat{t}_i \times \vec{r}^0_i)}{|\hat{t}_i \times \vec{r}^0_i|^2}, \label{eq:biot-savart}
\end{equation}
where $d\vec{\ell}$ denotes the vector arc measure on $\Gamma$, $\hat{t}_i$ is the unit tangent vector to the $i$th segment of $\Gamma$, $\vec{r}^0_i$ and $\vec{r}^1_i$ are, respectively, the vectors from the point $x$ to the initial and final vertices of segment $i$, and $\hat{r}^0_i$ and $\hat{r}^1_i$ are their normalized directions. In practice, we scale $\alpha$ by $10^{-3}$ to better match the normalization of our network weights; this only changes the mass minimization problem by a uniform scale. \Cref{fig:2d}(a) visualizes  a Biot-Savart field in 2D.

Enacting the choices above, we can use neural networks to solve the minimal mass problem:
\begin{equation}
\begin{aligned}
    &\argmin_\theta \|\dext f_\theta + \alpha_\Gamma \|_1 \\
    = &\argmin_\theta \E_{x \sim \U_U} \left[ \left|\nabla_x f_\theta(x) + \alpha^\sharp_\Gamma(x)\right| \right],
\end{aligned}  \label{eq:massmin-deep}
\end{equation}
where $f_\theta$ is a neural network with weights $\theta$, $\U_U$ is the uniform distribution over $U = [-1, 1]^d$. $\nabla$ is computed exactly via automatic differentiation, and $\alpha_\Gamma^\sharp$ is the boundary-dependent Biot-Savart field, computed in closed form.
The expectation in \eqref{eq:massmin-deep} is approximated by uniform sampling over $U$, yielding a method to compute minimal surfaces via stochastic gradient descent (SGD).

In relation to previous discretizations of currents and the minimal surface problem like \cite{wang2021computing}, we
\begin{enumerate*}[label={(\alph*)}]
\item represent $f$ via a neural network rather than a voxel grid;
\item evaluate $\alpha$ in closed-form; and
\item evaluate the mass norm as an expectation that is amenable to SGD.
\end{enumerate*}
These key choices allow our minimal surfaces to achieve arbitrary resolution.

\subsection{Modifying the Metric} \label{sec:metric}

Critical to the computer vision applications considered in this paper, we can 
use a background Riemannian metric to encode general surfaces that are not minimal under the Euclidean metric. The properties of mass norm minimization almost certainly carry over---in particular, the regularity of minima (see, e.g., \cite{morgan2003regularity}).

Let $g$ be a Riemannian metric given by
\begin{align}
    g(X,Y) = \langle AX,Y\rangle = \langle X,AY\rangle\quad\forall X,Y\in TU,
\end{align}
where $\langle \cdot,\cdot \rangle$ is the Euclidean inner product and
$A$ is a smoothly varying symmetric positive definite linear map on the tangent bundle (\(A_x\colon T_x U\rightarrow T_xU\)). The Riemannian pointwise norm for a \(k\)-form \(\zeta\) is given by
\begin{align}
    |\zeta_x|_g = |(A_x^{-\nicefrac{1}{2}})^*\zeta_x|,
\end{align}
where $| \cdot |$ is the Euclidean pointwise norm and $B^*\zeta$ denotes the pullback form defined by $B^*\zeta(X_1,\ldots,X_{k}) = \zeta(BX_1,\ldots,BX_k)$.  The $g$--mass norm is then:
\begin{align}
    \mathbf{M}_g([\omega])\!=\!\|\omega\|_{1,g}\!=\!\int_U |(A^{-\nicefrac{1}{2}})^*\omega| (\det A)^{\nicefrac{d}{2}}\dext\mathrm{vol}.
\end{align}
The differential equation \eqref{eq:bdry-de} and its solution \eqref{eq:alpha-hodge} are topological and do not change.

In summary, the Riemannian problem differs from the Euclidean one by a symmetric positive definite matrix $B_x$:
\begin{equation}
\begin{aligned}
        &\argmin_\theta \|\dext f_\theta + \alpha_\Gamma \|_{1, g} \\
        = &\argmin_\theta \E_{x \sim \U_U} \left[ \left|B_x(\nabla_x f_\theta(x) + \alpha^\sharp_\Gamma(x))\right|\right].
\end{aligned} \label{eq:massmin-deep-metric}
\end{equation}

\begin{figure}
    \centering
    \includegraphics[width=\linewidth]{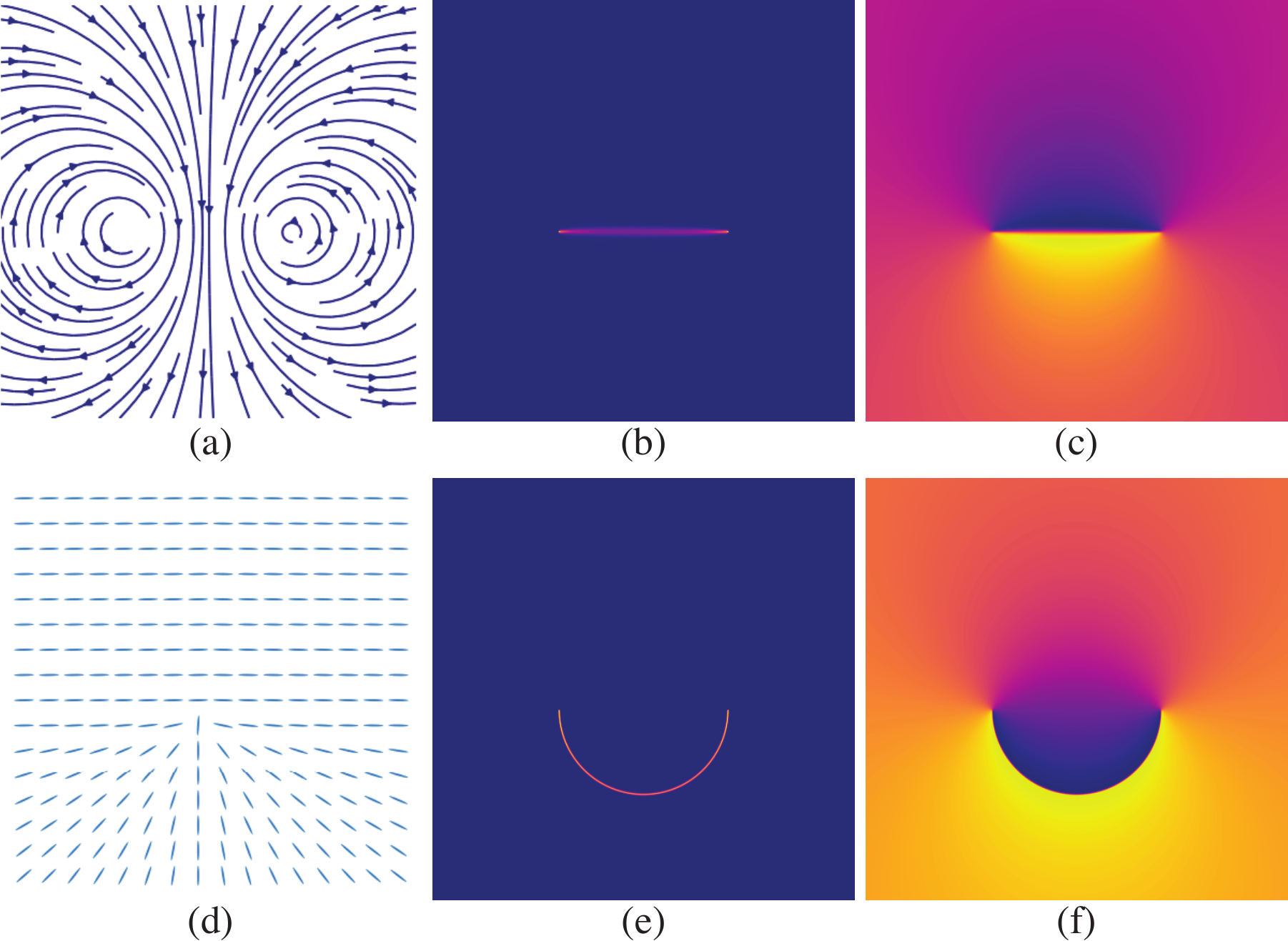}
    \caption{Minimizing the mass norm $\|df + \alpha\|_1$ under the Euclidean metric in two dimensions yields a line segment connecting the two boundary points (b). With our custom data-dependent background metric, we can reconstruct the semicircle as a current (e). $\alpha$ is shown as a vector field (a) and the custom metric is depicted by oriented ellipsoids (d, not to scale). Corresponding functions $f$ are shown at right (c and f).\vspace{-.15in}}
    \label{fig:2d}
\end{figure}

In the two-dimensional example depicted in \Cref{fig:2d}, minimizing the mass norm under the Euclidean metric yields a straight line segment (b). Changing the metric (d) yields a semicircle (e) instead. Corresponding density plots of $f$ are shown in (c) and (f), respectively.

\subsection{Loss Functions}
The main objective function optimized by our training procedures is the \emph{current loss}, which follows from \eqref{eq:massmin-deep-metric}:
\begin{equation}
    \LL_\textrm{curr}(\cdot) = \E_{x \sim \U_U} \left[ \left|B_x(\nabla_x f_\theta(x) + \alpha^\sharp_\Gamma(x))\right| \right].
\end{equation}
We approximate the expectation by a sample average over a sample drawn from the uniform distribution on $U$.

For minimal surface computation (\Cref{sec:minsurf}), we set $B_x =I$ for all $x \in U$. For surface reconstruction (\Cref{sec:reconstruction,sec:latent-learning}), we define
\begin{equation}
B_x = w_x(I - \hat{n}_{\proj_\Sigma(x)}\hat{n}_{\proj_\Sigma(x)}^\top), \label{eq:custom-metric}
\end{equation}
where $\Sigma$ is the ground truth surface, $\proj_\Sigma(x)$ is the closest point on $\Sigma$ to $x$, and $\hat{n}_{\proj_\Sigma(x)}$ is its unit normal. This positive semidefinite matrix, corresponding to a degenerate Riemannian metric, penalizes the current's deviation from agreement with the surface's orientation. A patch aligned with $\Sigma$ (i.e., where $\nabla f + \alpha_\Gamma^\sharp \parallel \hat{n}$) costs nothing.

When evaluating \Cref{eq:massmin-deep-metric}, for half of the samples in $U$, we set $w_x = 1$, and for the other half, we set
\begin{equation}
    w_x = \exp\left(- \frac{1}{2\sigma^2} \|x - \proj_\Gamma(x)\|^2_2\right),
\end{equation}
where $\proj_\Gamma(x)$ is the closest point on the boundary to $x$ under Euclidean distance, and $\sigma=0.1$ in practice. We find empirically that adding this boundary weighting, where samples close to the prescribed boundary have a higher contribution to the current loss, with a Gaussian falloff, slightly improves our learned surfaces, particularly near the boundary. See \Cref{sec:ablation} for an ablation study.

For surface reconstruction, we employ an additional loss term to guide our optimization. We define \emph{surface loss} as:
\begin{equation}
\LL_\textrm{surf}(\cdot)\!=\!\E_{x, \epsilon} \left[\left(\delta - f(x - \varepsilon n_x) + f(x + \varepsilon n_x)\right)^+\right],
\end{equation}
where $x \sim \U_\Sigma$, the uniform measure on the target surface $\Sigma$, $n_x$ is the (oriented) surface normal vector at a point $x \in \Sigma$, $\varepsilon$ and $\delta$ are small threshold. In practice, we set $\delta=0.01$, randomly pick $\varepsilon \sim \U_{[0.0199, 0.0201]}$, and approximate the expectation by sampling on $\Sigma$.

This hinge loss encourages the values of our learned function $f$ to differ by no less than a margin $\delta$ across the target surface. This objective may seem redundant as the metric \eqref{eq:custom-metric} already encourages alignment to the target surface. In fact, the two are complementary---the surface loss encourages $f$ to jump near the target surface, while the current loss ensures that the bandwidth of the jump decreases. We find that using the surface loss term helps our models converge to better optima (see our ablation study in \Cref{sec:ablation}).

\subsection{Network Architecture}
\label{sec:architecture}

\begin{figure}
    \centering
    \includegraphics[height=0.85\linewidth]{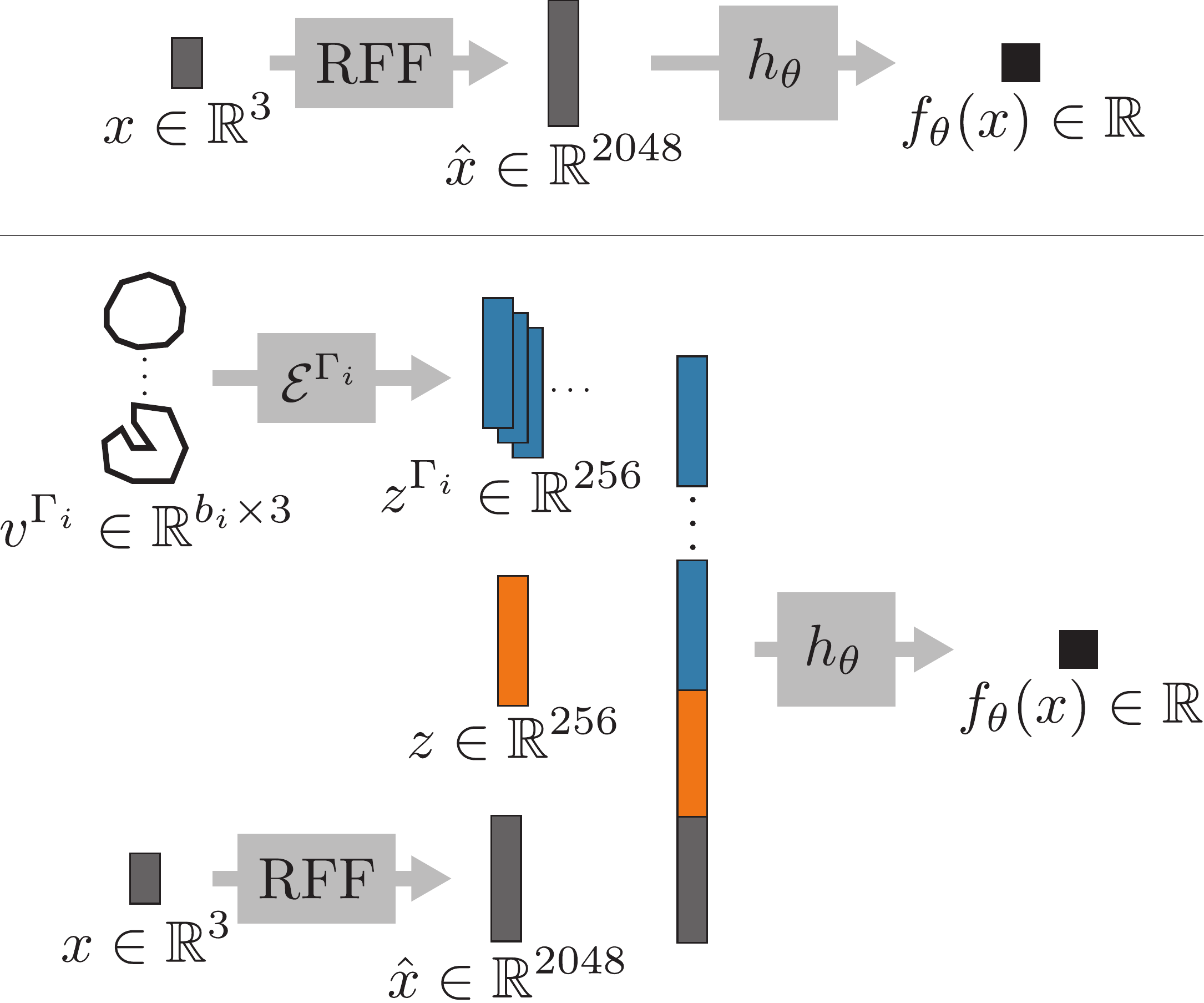}
    \caption{An overview of our network architectures for minimal surface optimization and single surface reconstruction (top) as well as shape space learning (bottom). An input point $x$ is first encoded using random Fourier features. These features are then optionally concatenated with latent codes corresponding to shape identity and boundary and finally decoded to a scalar output.\vspace{-.15in}}
    \label{fig:architecture}
\end{figure}

We learn a single current $\dext f_\theta + \alpha$, using a deep neural network to parameterize $f_\theta: \R^3 \to \R$. Given an input point $x \in \R^3$, we first project it onto a random Fourier feature (RFF) space, as in \cite{tancik2020fourfeat}, to obtain $\hat x \in \R^{2048}$. Our RFF coefficients are 2048-dimensional and sampled from $\N(0, 4)$. We then decode the RFF vector to a scalar value $f_\theta(x)$ using an MLP $h_\theta$, which consists of three hidden layers, each with 256 units and softplus nonlinearities. This pipeline is illustrated in the top half of \Cref{fig:architecture}.

Additionally, we propose a boundary-conditioned autodecoder architecture for learning families of currents (see \Cref{fig:architecture}, bottom). We initialize a latent code $z_j \sim \N(0, 0.1)$ for each mesh, and we encode the mesh boundary geometry using a boundary encoder $\mathcal E^\Gamma$. For shapes that have more than one boundary, we use a separate encoder for each boundary loop to obtain a set of boundary latent codes $z^{\Gamma_i}$. We then concatenate the latent codes along with the RFFs $[z_j \mid z_j^{\Gamma_1} \mid \ldots \mid z_j^{\Gamma_B} \mid \hat x]$ and pass this vector through a decoder $h_\theta$ as in the overfitting setting above.

Our boundary encoder inputs boundary vertices $v^\Gamma \in \R^{b \times 3}$. The encoder is a network with three 1-dimensional convolutional layers with stride 1 and circular boundary conditions. The first layer uses a kernel of size 5 while the latter two use kernels of size 3. Each layer has 256 channels, and we use ReLU after each layer except the last. After the convolutions, we take the mean across all the boundary vertices to obtain the boundary latent code. Circular convolutions combined with mean pooling ensure that our encoder is invariant to cyclic permutations of the vertices, which correspond to the same boundary geometry.

\section{Experimental Results}

We evaluate DeepCurrents experimentally by demonstrating results on minimal surface computation, overfitting for single surface reconstruction, and shape space learning and interpolation. We also show an ablation study to validate our main design choices. All of our models are trained on a single NVIDIA GeForce RTX 3090 GPU using Adam \cite{kingma2014adam}.

\subsection{Minimal Surfaces}
\label{sec:minsurf}

We use our method to compute minimal surfaces for three boundary configurations. We train each model for $10^5$ iterations ($\sim\!12$ minutes) with a learning rate of $0.0005$, sampling $\num{4096}$ points from the ambient space at each step and reducing the learning rate by a factor of $0.6$ every $\num{10000}$ steps. We only optimize $\mathcal{L}_\textrm{curr}$ with the Euclidean metric in these examples---we do not use $\mathcal{L}_\textrm{surf}$ or boundary weighting.

\begin{figure}
    \centering
    \includegraphics[height=0.85\linewidth]{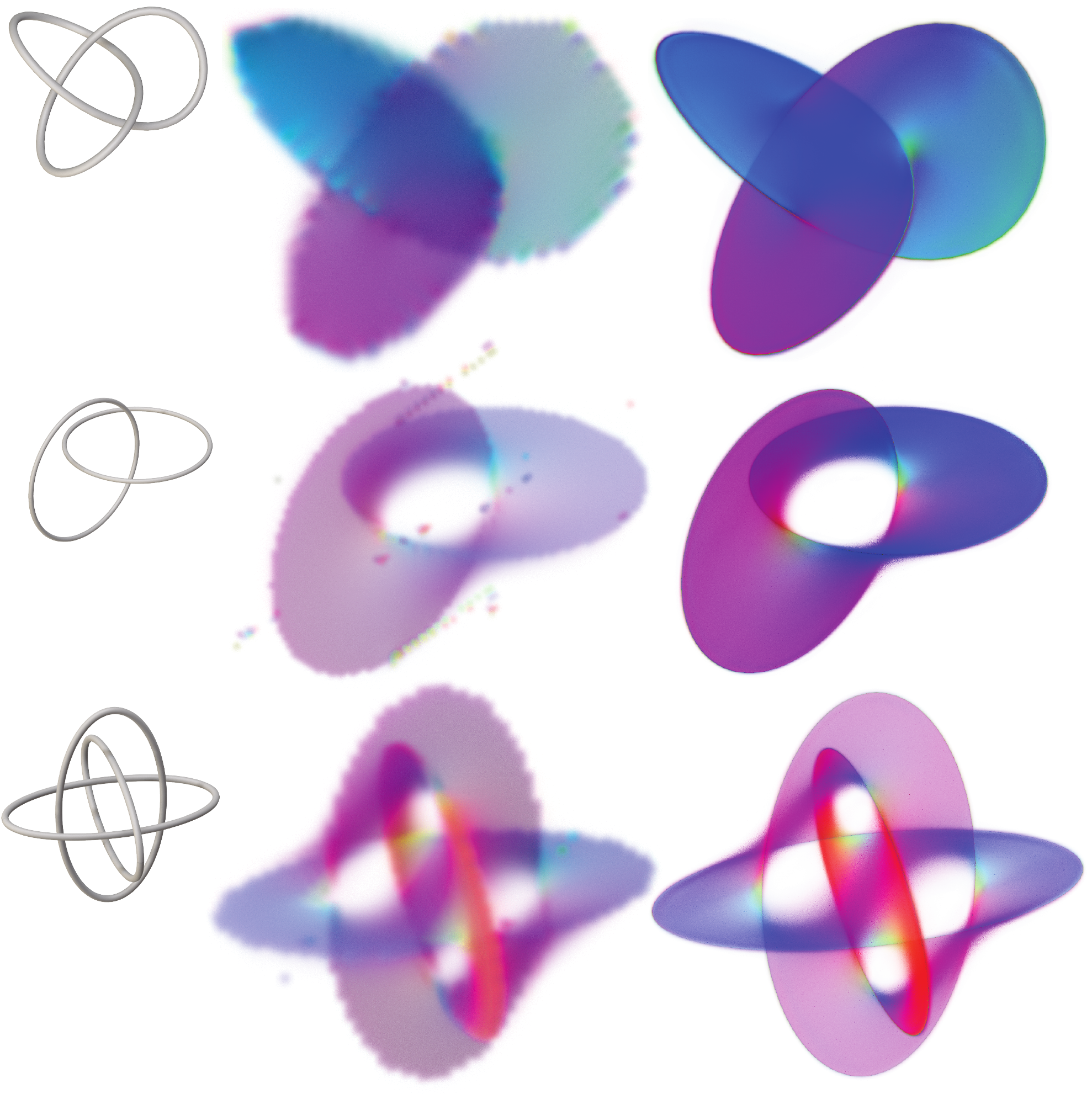}
    \caption{Minimal currents computed via \cite{wang2021computing} on a $90\!\times\!90\!\times\!90$ grid (middle) display prominent grid artifacts, especially near the boundary. In contrast, with a similar total number of parameters ($\num{725249}$ weights), our DeepCurrents achieve higher effective resolution (right). Boundaries (left) are the trefoil knot (top), Hopf link (middle), and Borromean rings (bottom).\vspace{-.15in}}
    \label{fig:minsurf-comparison}
\end{figure}

\Cref{fig:minsurf-comparison} compares our results to those from \cite{wang2021computing}, which uses a voxel grid; the colors represent local current orientation, which corresponds to the surface normal direction. For fair comparison, we choose the grid size to approximate our number of trainable parameters ($90^3 \approx \num{725249}$). While our learned currents adhere well to the smooth input boundaries, the currents of \cite{wang2021computing} show significant grid artifacts. Thus, our representation exhibits greater capacity to encode high-resolution surfaces with the same number of parameters

\subsection{Surface Reconstruction}
\label{sec:reconstruction}

We perform surface reconstruction using DeepCurrents by overfitting to several segmented parts of models from the FAUST human body dataset \cite{bogo2014faust}. We preprocess the data by splitting the mesh according to the provided segmentations, rigidly aligning all the models within each segmentation class, and rescaling them to fit into $[-0.5,0.5]^3$.

We train each model for $\num{10000}$ iterations ($\sim 4$ minutes) with an initial learning rate of $0.001$, decayed by a factor of $0.6$ every $\num{2000}$ iterations. We sample $\num{4000}$ random points from the ambient space (to compute $\mathcal{L}_\textrm{curr}$) and $\num{4000}$ points from the mesh surface (to compute $\mathcal{L}_\textrm{surf}$) at each step.

\begin{figure}
    \centering
    \includegraphics[width=\linewidth]{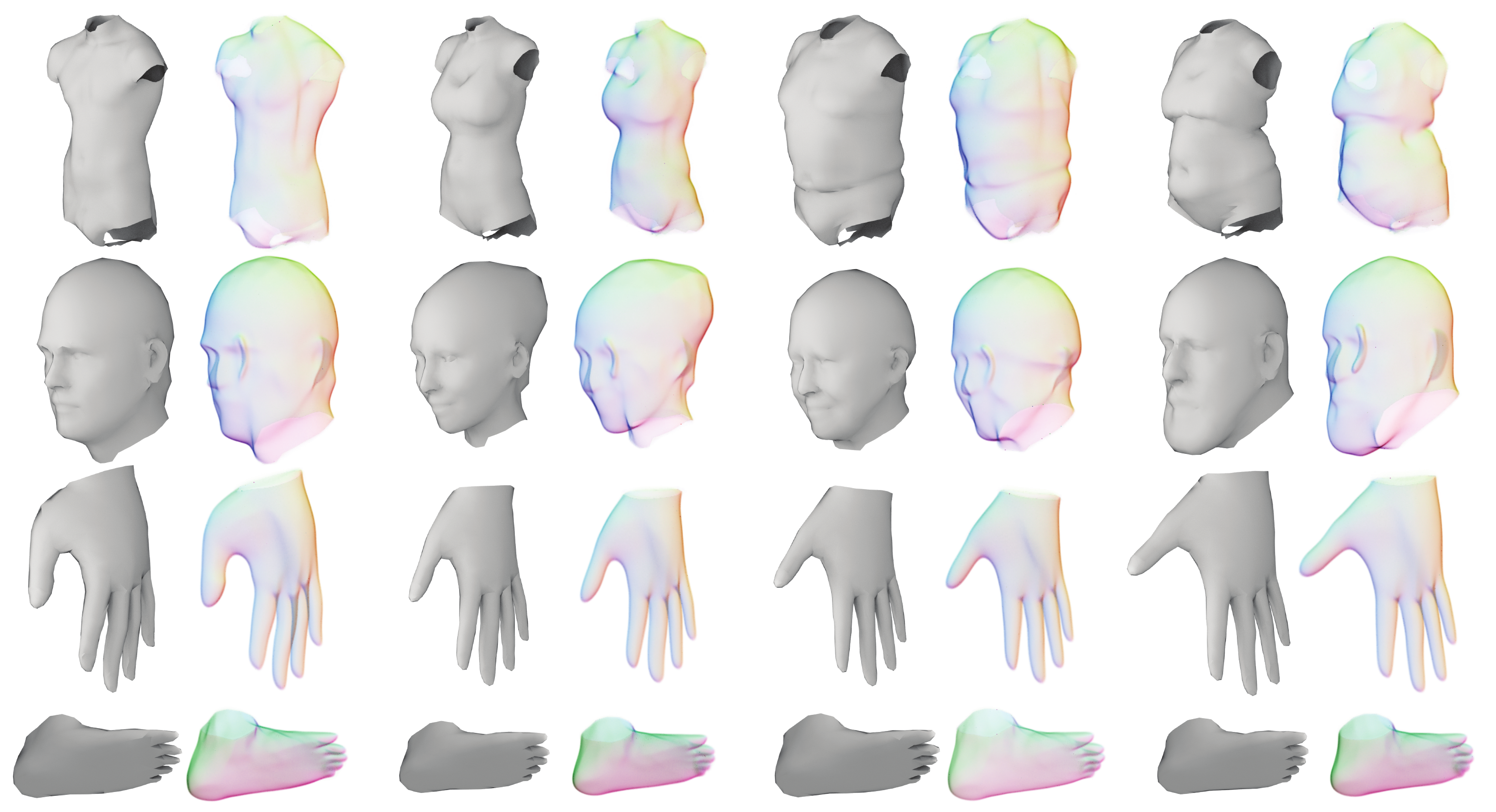}
    \caption{Human body surface reconstructions. We overfit DeepCurrents models to reconstruct several torso, head, hand, and foot meshes. We show a volume rendering of each learned current (right) next to the ground truth mesh (left).}
    \label{fig:reconstructions}
\end{figure}

We show results on torsos, heads, hands, and feet from randomly chosen models in \Cref{fig:reconstructions}. Our currents faithfully reconstruct the target geometry.

Additionally, we compare quantitatively to \cite{Venkatesh_2021_ICCV} in \Cref{tab:comparison}. We train their model for the same amount of time as ours on randomly picked models. Because their model predicts the closest point on the target surface given any input point, we use this to compute unidirectional Chamfer distance (i.e., $\E_{y \sim \U_\Sigma}[\operatorname{dist}_{\Sigma^*}(y)]$, where $\U_\Sigma$ is the uniform distribution on the ground truth mesh, and $\operatorname{dist}_{\Sigma^*}$ is Euclidean distance to the learned surface).

We do the same for our method by meshing our learned current: We compute the average value $s$ of $f$ over a boundary curve; this makes sense even if there are multiple boundary curves, assuming our surface has one connected component. Then, we extract a mesh of the level set $f^{-1}(s)$ using marching cubes. This level set is generically a closed surface containing our represented surface with boundary $\Sigma^*$ as a subset. We extract a mesh of $\Sigma^*$ by removing vertices $x$ for which $|\nabla_x f_\theta(x) + \alpha_\Gamma^\sharp(x)| < \delta$. In practice, we use $\delta=\num{5e-3}$.

Our method  consistently achieves better quality reconstructions than \cite{Venkatesh_2021_ICCV}.

\begin{table}
    \centering
    \begin{tabular}{ccc}
         \toprule
         Model & UCD (\cite{Venkatesh_2021_ICCV}) & UCD (ours)\\
         \midrule
         head & 0.0049 &\bf 0.0010 \\
         hand & 0.0045 & \bf 0.0011 \\
         torso & 0.0049 &\bf 0.00092\\
         foot & 0.0055 &\bf 0.00092\\
         \bottomrule
    \end{tabular}
    \caption{Quantitative comparison of unidirectional Chamfer distance to \cite{Venkatesh_2021_ICCV} on single surface reconstruction of random models from each of four shape categories.}
    \label{tab:comparison}
\end{table}

\subsection{Latent Space Learning}
\label{sec:latent-learning}

\begin{figure*}[ht]
    \centering
    \includegraphics[width=\linewidth]{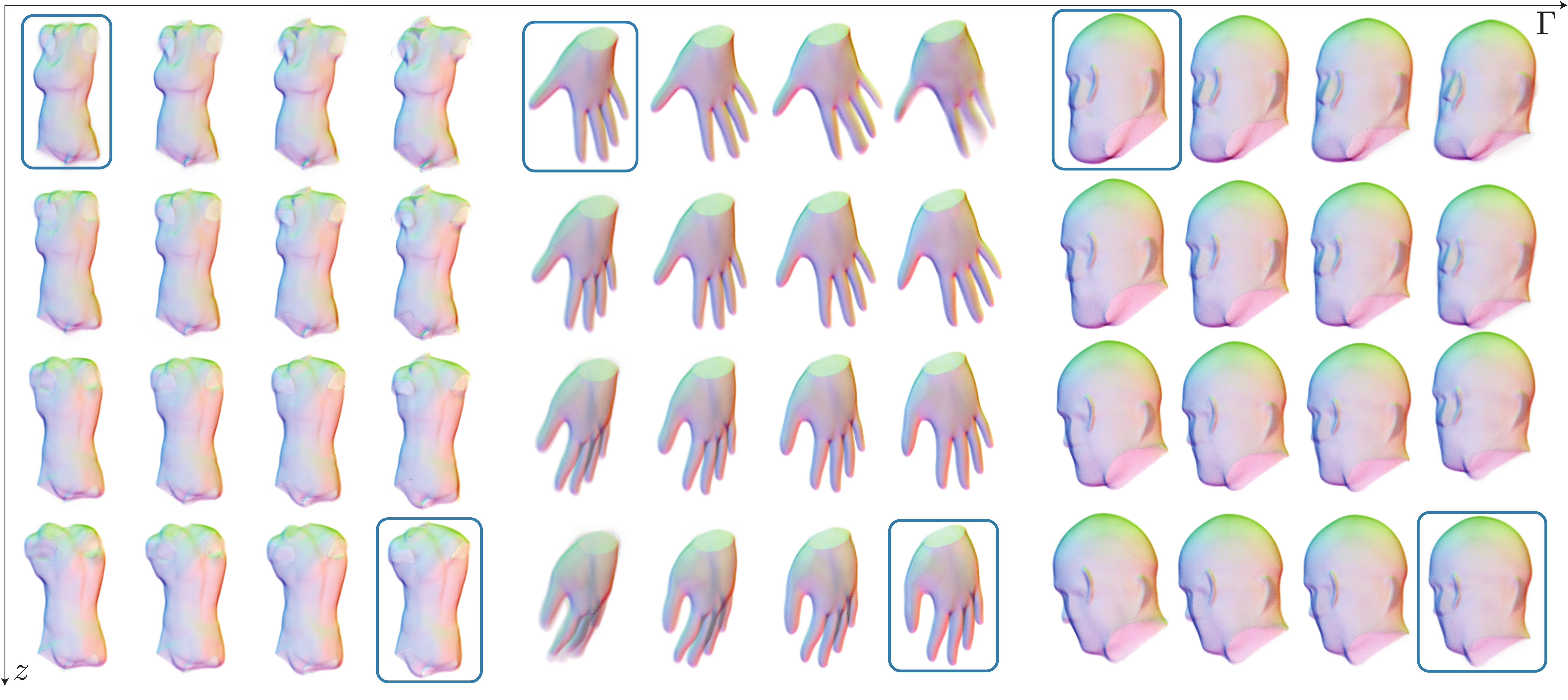}
    \caption{Interpolations of DeepCurrents in latent and boundary space. For each category, we pick two meshes from the training set (shown with a blue border) and interpolate linearly between the two boundaries (horizontal axis) as well as the two latent codes (vertical axis). The latent space interpolation yields a smooth transition between the two meshes while obeying the prescribed boundary interpolants.\vspace{-.15in}}
    \label{fig:interpolations}
\end{figure*}
We use our boundary-conditioned autodecoder (\Cref{sec:architecture}) to learn a disentangled representation that can interpolate in a high-dimensional learned latent space capturing shape identity while having explicit control over boundary geometry. We associate each mesh in our dataset with a random latent code (a trainable parameter), and, to disambiguate shape identity from boundary geometry, we perform random transformations. These transformations change the boundary shape while preserving the latent code.

At each iteration, we perform random augmentations to the target meshes: we rotate each mesh by sampling a value in $[-10^\circ, 10^\circ]$ for each Euler angle, we rescale each boundary loop of the mesh by a random factor between 0.85 and 1.15 along each of its two principal directions, we propagate these transformations to the entire mesh using harmonic skinning weights, and we shift the mesh by a random offset between $-0.05$ and $0.05$ in each dimension.

We train a model for each shape category for $\num{300000}$ iterations (about 10 hours) with an initial learning rate of $0.0004$, decayed by a factor of $0.5$ every $\num{60000}$ iterations. At each step, we sample a random batch of 8 meshes and sample $\num{4000}$ points from each mesh.

In Figure~\ref{fig:interpolations}, we pick two models from each shape category and independently interpolate between their boundaries and latent identities. Our model disentangles high-level pose and style while respecting the prescribed geometry.

\subsection{Ablation Study} 
\label{sec:ablation}

\begin{figure}
    \centering
    \includegraphics[width=\linewidth]{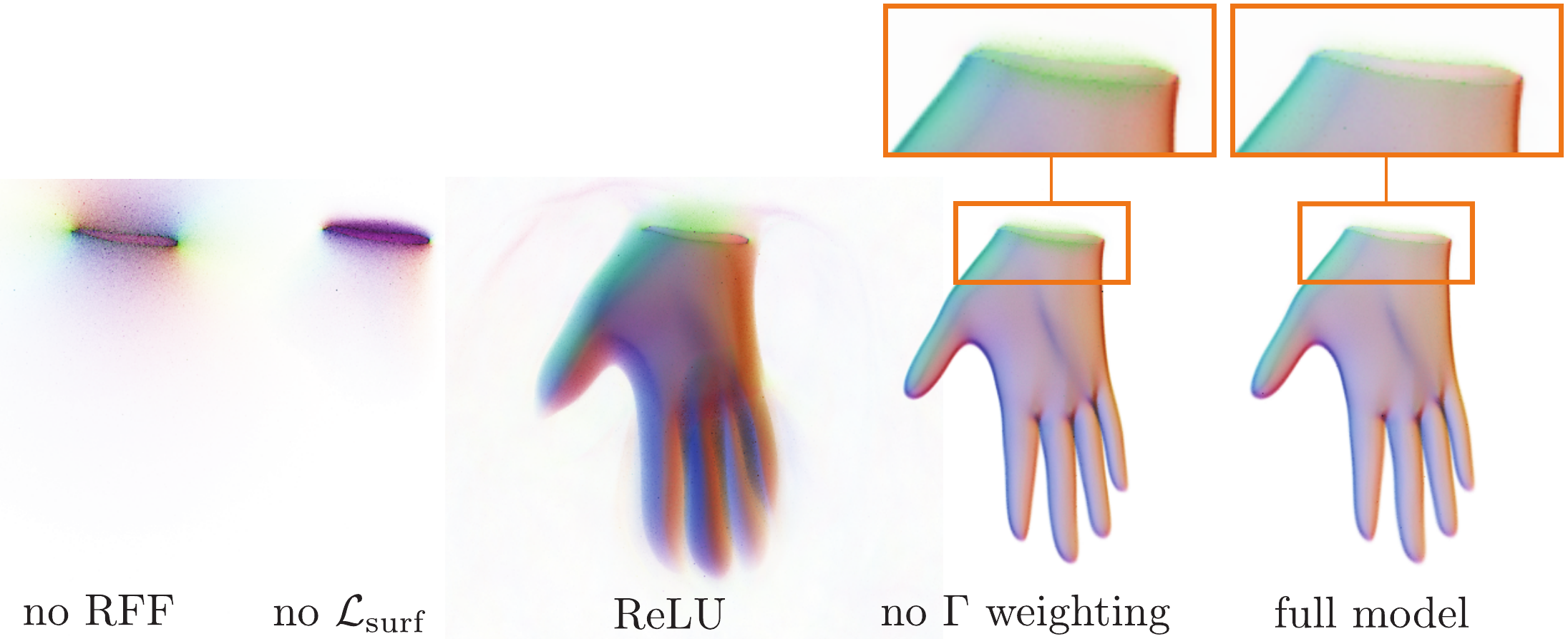}
    \caption{Ablation study. From left to right: reconstruction results on a hand without random Fourier features, without our surface loss term, without boundary weighting, and our full model.\vspace{-.15in}}
    \label{fig:ablation}
\end{figure}

We validate some of our key design choices. In \Cref{fig:ablation}, we overfit five models to the same hand mesh. While our full model achieves a sharp reconstruction, removing boundary weighting from our current loss metric yields a fuzzier surface around the boundary. Changing the softplus activation functions in $h_\theta$ to ReLUs makes the entire learned surface significantly less sharp, which we conjecture is due to ReLU's zero second derivative when optimizing currents $\dext f + \alpha$. Removing the surface loss term from our optimization fails to recover much of the target surface, supporting our claim that surface loss significantly helps convergence. Finally, foregoing the projection of the input points onto random Fourier features prevents the model from learning.

\section{Conclusion}

By adopting tools from geometric measure theory, we have constructed a neural implicit representation for surfaces with boundary. Our SGD approach to mass norm minimization enables computing minimal surfaces with arbitrary resolution, in contrast to previous work that represents currents on a fixed-resolution grid. In addition, by constructing a background metric, we can engineer a mass minimization problem to encode an arbitrary surface. Combining this construction with the expressive power of neural representations, we can encode whole families of surfaces.%, enabling interpolation and other applications.

We see DeepCurrents as a key tool for building flexible neural surface representations. Stitching together DeepCurrents along their boundaries would produce a hybrid surface representation where the explicit boundary curves provide ``handles'' for user control. Such a representation would be applicable where the target surface is decomposed into parts. Unlike, say, a mesh decomposition, a DeepCurrent decomposition would not require the parts to have simple shapes or even to be simply connected.

Another direction for future work would be to investigate other loss functions and optimization problems that can be expressed in the language of currents. For example, the convex problems studied in \cite{Mollenhoff:2019:LVV} could be optimized using a neural representation and SGD. One could also compute minimal currents in spaces such as the rotation groups $\mathrm{SO}(d)$ or special Euclidean groups $\mathrm{SE}(d)$. Mass minimization in this context could provide a useful prior for reconstruction of shapes that come with an orientation or frame field, or it could exploit the Gauss map to encode a smoothness prior.

Another extension of our method would be to support periodic minimal surfaces, i.e., replacing the domain $[-1, 1]^3$ by the torus $\mathbb{T}^3$. This would require a modification of our explicit $\alpha$ and the evaluation of $\dext f$ at the boundary.

While our latent space model often produces high-quality interpolants, they are not explicitly regularized to encourage them to look like surfaces. This sometimes yields fuzzy results (see \Cref{fig:interpolations}, top right among the hands). Future work could design loss terms to ensure interpolants remain minimal with respect to some metric---analogously to the eikonal regularization for SDFs in \cite{amos2020}.
\section*{Acknowledgements}
The MIT Geometric Data Processing group acknowledges the generous support of Army Research Office grants W911NF2010168 and W911NF2110293, of Air Force Office of Scientific Research award FA9550-19-1-031, of National Science Foundation grants IIS-1838071 and CHS-1955697, from the CSAIL Systems that Learn program, from the MIT--IBM Watson AI Laboratory, from the Toyota--CSAIL Joint Research Center, from a gift from Adobe Systems, from an MIT.nano Immersion Lab/NCSOFT Gaming Program seed grant, and from the Skoltech--MIT Next Generation Program. This work was also supported by the National Science Foundation Graduate Research Fellowship under Grant No.\ 1122374. David Palmer appreciates the generous support of the Hertz Fellowship and the MathWorks Fellowship.

{\small
\bibliographystyle{abbrvnat}
\bibliography{deepcurrents}
}

\end{document}